\documentclass{article}
\usepackage{spconf}
\usepackage{amsmath}
\usepackage{graphicx}
\usepackage{amsthm}
\usepackage{amsfonts}
\usepackage{float}
\usepackage{xspace}
\usepackage{bm}
\usepackage{color}
\usepackage{tabu}
\usepackage{CJK}
\usepackage{booktabs}
\usepackage{xcolor}
\usepackage{multirow}
\usepackage{array}
\usepackage{graphicx}
\usepackage{url}
\usepackage{makecell}
\usepackage{subcaption}

\title{Improving Stability in Simultaneous Speech Translation: A Revision-Controllable Decoding Approach}
\name{Junkun Chen, Jian Xue, Peidong Wang, Jing Pan, Jinyu Li}
\address{Microsoft}
\begin{document}
%\ninept
%
\copyrightnotice{979-8-3503-0689-7/23/\$31.00~\copyright2023 IEEE}
\maketitle

\begin{abstract}

Simultaneous Speech-to-Text translation serves a critical role in real-time crosslingual communication. 
Despite the advancements in recent years, 
challenges remain in achieving stability in the translation process, 
a concern primarily manifested in the flickering of partial results.
In this paper, we propose a novel revision-controllable method designed to address this issue. 
Our method introduces an allowed revision window within the beam search pruning process to
screen out candidate translations likely to cause extensive revisions,
leading to a substantial reduction in flickering 
and, crucially,
providing the capability to completely eliminate flickering.
The experiments demonstrate the proposed method can significantly improve the decoding stability
without compromising substantially on the translation quality.

\end{abstract}
\begin{CJK*}{UTF8}{gbsn}

% !TEX root = main.tex

\newcommand{\tuple}[1]{\ensuremath{\langle {#1} \rangle}}

\newcommand{\featuresum}[1]{\ensuremath{\sum_i \lambda_i f_i({#1})}}

\newcommand{\fvec}[1]{\ensuremath{\vec{f} ({#1})}}

\newcommand{\fiof}[1]{\ensuremath{f_i ({#1})}}

\newcommand{\fofi}[0]{\ensuremath{f_i}}

\newcommand{\featureprod}[1]
{\ensuremath{\vec{\lambda} \cdot \fvec{#1} } }

\newcommand{\efprod}[1]{\ensuremath{e^{\featureprod{#1}}}}
\newcommand{\efsum} [1]{\ensuremath{e^{\featuresum {#1}}}}

\newcommand{\lami}{\ensuremath{\lambda_i}}

\newcommand{\empExp}[1]{\ensuremath{\mbox{\~{E}} [{#1}]}}
\newcommand{\expect}{\ensuremath{\mathbb{E}}}

\newcommand{\ks}{\ensuremath{k}}
\newcommand{\Ks}{\ensuremath{K}}
\newcommand{\kbest}{\ks-best\xspace}
\newcommand{\topk}{top-\ks\xspace}
\newcommand{\kBest}{\Ks-Best\xspace}

\newcommand{\notes}[1]{}%{\it {\small {#1}}}}

\newcommand{\acite}[1]{({--#1})}

\newcommand{\define}[1]{{\bf Definition.} {\em {#1}}}
\newcommand{\zerobar}{\ensuremath{\overline{0}}\xspace}
\newcommand{\onebar}{\ensuremath{\overline{1}}\xspace}

\newcommand{\Items}{\ensuremath{\mathcal{I}}\xspace}

\newcommand{\oplusk}{\ensuremath{\oplus_{\tiny k}}}
\newcommand{\otimesk}{\ensuremath{\otimes_{\tiny k}}}

\newcommand{\zerok}{\ensuremath{\overline{0}^k}\xspace}
\newcommand{\onek}{\ensuremath{\overline{1}^k}\xspace}

\newcommand{\scalar}{\ensuremath{\odot_k}}

\newcommand{\kdim}{\ks-dimensional\ }

% \newlistof{defin}{def}{List of Definitions}

% \newcommand{\defin}[1]{%
% \refstepcounter{defin}
% \par\noindent\textbf{Definition \thedefin. #1}
% \addcontentsline{ans}{defin}{\protect\numberline{\thedefin}#1}\par}

% for amsthm
\theoremstyle{definition}
\newtheorem{definition}{Definition}
\theoremstyle{plain}
\newtheorem{theorem}{Theorem}
\newtheorem{lemma}{Lemma}

\newcommand{\gc}{\ensuremath{|G|}\xspace}
\newcommand{\ntsize}{\ensuremath{|N|}\xspace}
\newcommand{\nt}{\ensuremath{N}\xspace}
\newcommand{\rulepernt}{\ensuremath{R}\xspace}

\newcommand{\Ttimesk}{\ensuremath{T_{\otimesk}}\xspace}
\newcommand{\Tplusk}{\ensuremath{T_{\oplusk}}\xspace}
\newcommand{\maxk}{\ensuremath{\mbox{\bf max}_k}\xspace}

\newcommand{\tenexp}[1]{\ensuremath{10^{{-#1}}}\xspace}

\newcommand{\naive}{na\"{\i}ve\xspace}

\newcommand{\perc}[1]{\ensuremath{{#1}\%}\xspace}

\newcommand{\veca}{\ensuremath{\mathbf{a}}}
\newcommand{\vecalpha}{\ensuremath{\mathbf{\alpha}}}
\newcommand{\vecb}{\ensuremath{\mathbf{b}}}
\newcommand{\vecp}{\ensuremath{\mathbf{p}}}
\newcommand{\klogk}{\ensuremath{k\log k}}
\newcommand{\vecz}{\ensuremath{\mathbf{z}}}
\newcommand{\vech}{\ensuremath{\bm{h}}\xspace}
\newcommand{\vecW}{\ensuremath{\mathbf{W}}}
\newcommand{\vecS}{\ensuremath{\mathbf{S}}}
\newcommand{\vecs}{\ensuremath{\mathbf{s}}\xspace}
\newcommand{\better}{\ensuremath{\preceq}}

\newcommand\dbar[1]{\overline{\overline{#1}}}

\newcommand{\opt}{\ensuremath{\mbox{{\bf min}}_\better}}
\newcommand{\mergek}{\ensuremath{\mbox{{\bf merge}}_{\better k}}\xspace}
\newcommand{\multk}{\ensuremath{\mbox{{\bf mult}}_{\better k}}\xspace}

\newcommand{\kw}{\ensuremath{\mathbf{w}}}

\newcommand{\kwi}{\ensuremath{w_i}}
\newcommand{\wejv}[3]{\ensuremath{w^{#1}_{{#2}}({#3})}}

\newcommand{\vecj}{\ensuremath{\mathbf{j}}}
\newcommand{\vecC}{\ensuremath{\mathbf{C}}}
\newcommand{\vecD}{\ensuremath{\mathbf{D}}}
\newcommand{\vecu}{\ensuremath{\mathbf{u}}\xspace}
\newcommand{\vecU}{\ensuremath{\mathbf{U}}}
\newcommand{\vecjhat}{\ensuremath{\hat{\mathbf{j}}}}
\newcommand{\vecone}{\ensuremath{\mathbf{1}}}
\newcommand{\dbp}[2]{\ensuremath{\langle{#1}, {#2}\rangle}}

\newcommand{\reviewtimetoday}[3]{
    \AtBeginDocument{
  }}

% for submission
\iffalse
\renewcommand{\marginpar}[1]{}
\fi

%\newcommand{\comment}[1]{\marginpar{\raggedright{\em{\small #1}}}}

\newcommand{\ith}[1]{\ensuremath{i^{{th}}}}
\def\alglazy{Algorithm 3}

\newcommand{\JnM}{Jim\'enez and Marzal\xspace}

\newcommand{\srule}[1]{\ensuremath{\mathrm{#1}}}
\newcommand{\goesto}{\ensuremath{\rightarrow}\xspace}
\newcommand{\chn}[1]{\mbox{{\it {#1}}}}

\newcommand{\ngram}{\ensuremath{\text{$n$-gram}}\xspace}
\newcommand{\ngrams}{\ensuremath{\text{$n$-grams}}\xspace}

\newcommand{\ngramitem}[5]{
\ensuremath{
\left(
\begin{smallmatrix}
\mbox{\small {#4}} &\cdots & \mbox{\small {#5}} \\
&
%\!\!_{#2}
\mbox{#1}
%_{#3}
 &
\\ {#2} & & {#3}
\end{smallmatrix}
 \right)
}}

\newcommand{\specialngramitem}[7]{
\begin{math}
\left(
\begin{smallmatrix}
\mbox{\small {#2}}\ \cdots\ \mbox{\small{#3}}
& \cdots &
\mbox{\small {#4}}\ \cdots\ \mbox{\small {#5}} \\
& \mbox{{#1}} & \\
{#6} & & {#7}
\end{smallmatrix}
 \right)
\end{math}
}

\newcommand{\chartitem}[3]{
\ensuremath{(\mbox{#1},{#2},{#3})}
}

\newcommand{\vtnt}[1]{\ensuremath{V_{\mbox{\tiny #1}}}}
%%% \bigram{a}{b} means (a,b) is a bigram pair. P (b | a)!
\newcommand{\bigram}[2]{\ensuremath{\Pr(\mbox{\small #2} \mid \mbox{\small #1})}}

\newcount\permx
\newcount\permy
\def\permdot#1#2{
\permx=#1 \advance\permx by-1
\permy=#2 \advance\permy by-1
\psframe[fillcolor=black, fillstyle=solid]
(\permx,\permy)(#1, #2)
}

%%% note: realcalc.sty has a fatal bug : 23-0.5=23.5.
%%% so i have to do this... +1-0.5 thing
\newcommand{\minushalf}[1]{
    \Radd{\aaa}{#1}{-1}
    \Radd{\aaa}{\aaa}{0.5}
    \Rtrunc{\aaa}{1}{\aaa}
    \aaa
}

\newcommand{\permnt}[3]{
\rput({#1},{#2}){\huge\color{white} {#3}}
}

\newcommand\union{\cup}
\newcommand\intersect{\cap}

\newcommand\veczero{\ensuremath{\mathbf{0}}}
\newcommand\vecq{\ensuremath{\mathbf{q}}}
\newcommand\vecn{\ensuremath{\mathbf{n}}}
\newcommand{\argmax}{\operatornamewithlimits{\mathbf{argmax}}}
\newcommand{\argtop}{\operatornamewithlimits{\mathbf{argtop}}}
\newcommand{\toptop}{\operatornamewithlimits{\mathbf{top}}}
\newcommand{\argmin}{\operatornamewithlimits{\mathbf{argmin}}}

\newcommand{\deltahat}{\ensuremath{\hat{\delta}}\xspace}

\newcommand{\xrs}{{\bf xRs}\xspace}
\newcommand{\xrsln}{\mbox{1-{\bf xRLNs}}\xspace}
\newcommand{\RLN}{{\bf RLN}\xspace}

\newcommand{\foot}[1]{\ensuremath{^{\tiny({#1})}_{\downarrow}}}

\newcommand{\chnprd}{{\tiny \ensuremath{_{\circ}}}}

\newcommand{\ckyitem}[3]{\ensuremath{{\mbox{#1}}_{{{#2},\,{#3}}}}\xspace}
\newcommand{\treeitem}[2]{\ensuremath{{\mbox{#1}}_{#2}}\xspace}
\newcommand{\nodent}[3]{\ensuremath{\treeitem{#1}{#2}:{#3}}}

\newcommand{\coverage}[1]{\ensuremath{(\mbox{{#1}})}\xspace}
\newcommand{\lmcov}[2]{\ensuremath{(\mbox{#1}, ^{\mbox{\scriptsize #2}}\!)}\xspace}
\newcommand{\phritem}[3]{\ensuremath{\coverage{#1}:({#2}, \mbox{\footnotesize ``#3''})}\xspace}
\newcommand{\lmphritem}[4]{\ensuremath{\lmcov{#1}{#2}:({#3}, \mbox{\footnotesize ``#4''})}\xspace}
\newcommand{\phrpair}[2]{\ensuremath{\langle \mbox{\chn{#1}, {#2}}\rangle}\xspace}

\newcommand{\lmitem}[3]{\ensuremath{({{#1}}^{#2 \star #3})}\xspace}
\newcommand{\lmckyitem}[5]{\ensuremath{(\mbox{#1}_{\mbox{\scriptsize\ ({#2},{#3})}}^{\scriptsize\ \mbox{#4}\ \star\ \mbox{#5}})}\xspace}

\def\wlm{$\mathord+\textrm{LM}$\xspace}
\def\wolm{$\mathord-\textrm{LM}$\xspace}

\newcommand{\sym}[1]{\textrm{#1}\xspace}
\newcommand{\VP}{\sym{VP}}
\newcommand{\PP}{\sym{PP}}

\newcommand{\startsym}{\mbox{\scriptsize \texttt{<s>}}\xspace}
\newcommand{\stopsym}{\mbox{\scriptsize \texttt{</s>}}\xspace}
\newcommand{\TOP}{\sym{TOP}}
\newcommand{\plm}[2]{\ensuremath{P_{lm}(\mbox{#2}\mid\mbox{#1})}}

\newcommand{\order}{\ensuremath{\mathcal{O}}}

\newcommand{\wocombo}{\ensuremath{h}\xspace}
\newcommand{\mincombo}{\ensuremath{h_{\mathit{combo}}}\xspace}
\newcommand{\tb}{\ensuremath{{\mathit{TB}}}\xspace}

\newcommand{\hyp}[1]{\mbox{\tiny ``{#1}''}}
\newcommand{\hl}{\ensuremath{k}\xspace}

\newcommand{\cand}{\ensuremath{\mathit{cand}}\xspace}
\newcommand{\buf}{\ensuremath{\mathit{buf}}\xspace}
\newcommand{\bound}{\ensuremath{\mathit{bound}}\xspace}

\newcommand{\lazyjbest}{\ensuremath{\textproc{LazyJthBest}}\xspace}

\newcommand{\backwardstar}{\ensuremath{\mathit{IN}}\xspace}

\newtheorem{proposition}[theorem]{Proposition}

\newcommand{\ff}{\ensuremath{\mathbf{f}}\xspace}
\newcommand{\ee}{\ensuremath{\mathbf{e}}\xspace}
\newcommand{\al}{\ensuremath{\mathbf{a}}\xspace}

\newcommand{\pt}{\ensuremath{p_{\mbox{t}}}}
\newcommand{\pd}{\ensuremath{p_{\mbox{d}}}}

\newcommand{\current}{\color{blue}{\fbox{{\bf C}}}\xspace}
\newcommand{\future}{\color{red}{\fbox{{\bf F}}}\xspace}

\newcommand{\boxnum}[1]{{\setlength{\fboxsep}{1pt}\raisebox{1pt}{\hspace{1pt}\fbox{\tiny #1}\hspace{1pt}}}}
\newcommand{\ind}[1]{\ensuremath{_{\kern-0.5pt\boxnum{#1}}}}
\newcommand{\BS}{\ensuremath{\mathit{IN}}\xspace}
\newcommand{\NTX}{\text{X}}
\newcommand{\NTS}{\text{S}}

\def\FS{\mathit{FS}\xspace}
\def\BS{\mathit{BS}\xspace}
\def\bfR{\mathbf{R}\xspace}

\newcommand{\head}{\ensuremath{\mathit{head}}\xspace}
\newcommand{\tails}{\ensuremath{\mathit{tails}}\xspace}

\newcommand{\Deriv}{\ensuremath{\mathscr{D}}\xspace}
\newcommand{\hhat}{\ensuremath{\hat{h}}\xspace}

\newcommand{\opluseq}{\ensuremath{\ \oplus=}\xspace}

\newcommand{\lianheguo}{\chn{Li\'anh\'egu\'o}\xspace}
\newcommand{\jiandu}{\chn{ji\=and\=u}\xspace}

\newcommand{\chnBushi}{\chn{B\`ush\'i}\xspace}
\newcommand{\chnyu}{\chn{y\v{u}}\xspace}
\newcommand{\chnle}{\chn{le}\xspace}
\newcommand{\chnShalong}{\chn{Sh\=al\'ong}\xspace}
\newcommand{\chnBaoweier}{\chn{B\`aow\=eier}\xspace}
\newcommand{\chnjuxing}{\chn{j\v{u}x\'ing}\xspace}
\newcommand{\chnhuitan}{\chn{hu\`it\'an}\xspace}

\newcommand{\mybullet}{\ensuremath{\bullet\hspace{-0.05mm}}}
\newcommand{\myunderscore}{\ensuremath{\mbox{\Large\_}}}

\newcommand{\nocov}{$_0$\myunderscore\myunderscore\myunderscore\myunderscore\myunderscore\myunderscore}
\newcommand{\onecov}{\mybullet$_1$\myunderscore\myunderscore\myunderscore\myunderscore\myunderscore}
\newcommand{\halfcov}{\mybullet\myunderscore\myunderscore\mybullet\mybullet\mybullet$_{6}$}
\newcommand{\fullcov}{\mybullet\mybullet\mybullet$_{3}$\mybullet\mybullet\mybullet}

\newcommand{\fullncov}{\mybullet\mybullet\ldots\mybullet}

\long\def\signature#1{%
% \begin{flushleft}
\begin{center}
% \begin{minipage}{6in}
\parindent=0pt
\shortstack{\vrule width 3in height 0.4pt\\ #1}
% \end{minipage}
\end{center}
% \end{flushleft}
}

% forest rerank acl 2008
\newcommand{\feat}[1]{{\bf {#1}}}
\newcommand{\Gen}[1]{\ensuremath{\mathit{cand}({#1})}\xspace}
\newcommand{\yhat}{\ensuremath{\hat{y}}\xspace}
\newcommand{\ystar}{\ensuremath{y^*}\xspace}
\newcommand{\yplus}{\ensuremath{y^+}\xspace}
\newcommand{\cost}{\ensuremath{\mathit{c}}\xspace}
\newcommand{\oracle}{\ensuremath{\mathit{oracle}}\xspace}
\newcommand{\ora}{\ensuremath{\mathit{ora}}\xspace}
\newcommand{\cnj}{charniak+johnson:2005}
\newcommand{\coll}{collins:2000}
\newcommand{\dom}{\ensuremath{\mathit{dom}}\xspace}
\newcommand{\gold}{\ensuremath{\mathit{gold}}\xspace}

\newcommand{\myoval}[1]{\ovalnode[linestyle=none,fillstyle=solid,fillcolor=lightgray]{foo}{#1}}
\newcommand{\mybox}[1]{\psframebox[linestyle=none,linewidth=0pt,fillstyle=solid,fillcolor=lightgray]{#1}}

\newcommand{\vecx}{\ensuremath{{\mathbf{x}}}\xspace}
\newcommand{\vecy}{\ensuremath{{\mathbf{y}}}\xspace}

\newcommand{\xbar}{\ensuremath{\overline{x}}\xspace}
\newcommand{\wbar}{\ensuremath{\overline{w}}\xspace}
\newcommand{\xbarbar}{\ensuremath{\dbar{x}}\xspace}
\newcommand{\wbarbar}{\ensuremath{\dbar{w}}\xspace}

\newcommand{\vecybar}{\ensuremath{\overline{\vecy}}\xspace}

\newcommand{\vecw}{\ensuremath{\bm{w}}\xspace}
\newcommand{\vece}{\ensuremath{\bm{e}}\xspace}
\newcommand{\vecc}{\ensuremath{\mathbf{c}}\xspace}
\newcommand{\vecd}{\ensuremath{\mathbf{d}}\xspace}
\newcommand{\vecR}{\ensuremath{\mathbb{R}}\xspace}
\newcommand{\vecr}{\ensuremath{\mathbb{r}}\xspace}
\newcommand{\vectheta}{\ensuremath{\mathbf{\theta}}\xspace}
\newcommand{\vecmu}{\ensuremath{\mathbb{\mu}}\xspace}

\newcommand{\vecv}{\ensuremath{\mathbf{v}}\xspace}
\newcommand{\vecf}{\ensuremath{\mathbf{f}}\xspace}
\newcommand{\vecfl}{\ensuremath{\mathbf{f}_L}\xspace}

\newcommand{\heap}{\ensuremath{\mathit{heap}}\xspace}

\newcommand{\feature}[1]{$\langle$ {#1} $\rangle$\xspace}

\newcommand{\defineq}{\ensuremath{\; \triangleq\; }\xspace}
\newcommand{\extradgt}{\color{white}{0}\xspace}

\newcommand{\funit}{\ensuremath{\mathring{f}}\xspace}
\newcommand{\vecfunit}{\ensuremath{\mathring{\vecf}}\xspace}
\newcommand{\Done}[1]{\ensuremath{D_1(#1)}}

% kbest paper 2005
\let\algsize\normalsize
\newcommand{\algline}[2]{%
\begin{center}
\algsize$\text{#1:}\hspace{1em}#2$
\end{center}}

\def\Dhat{\hat{D}}
\def\vecDhat{\mathbf{\hat{D}}}

%%%%%%%%%%%%%%% pinyins

\newcommand{\Baoweier}{\chn{B\`aow\=eier}\xspace}
\newcommand{\Bushi}{\chn{B\`ush\'i}\xspace}
\newcommand{\bushi}{\Bushi}
\newcommand{\Shalong}{\chn{Sh\=al\'ong}\xspace}
\newcommand{\yu}{\chn{y\v{u}}\xspace}
\newcommand{\hai}{\chn{h\'ai}\xspace}
\newcommand{\jiang}{\chn{ji\=ang}\xspace}
\newcommand{\huiwu}{\chn{hu\`iw\`u}\xspace}
\newcommand{\jinyibu}{\chn{j\`iny\={\i}b\`u}\xspace}
\newcommand{\juxing}{\chn{j\v{u}x\'ing}\xspace}
\newcommand{\huitan}{\chn{hu\`it\'an}\xspace}
\newcommand{\dangtian}{\chn{d\=angti\=an}\xspace}
\newcommand{\jiu}{\chn{ji\`u}\xspace}
\newcommand{\zhongdong}{\chn{zh\=ongd\=ong}\xspace}
\newcommand{\weiji}{\chn{w\=eij\=\i}\xspace}
\newcommand{\fuze}{\chn{f\`uz\'e}\xspace}

\newcommand{\zongtong}{\chn{z\v{o}ngt\v{o}ng}\xspace}
\newcommand{\zai}{\chn{z\`ai}\xspace}
\newcommand{\mosike}{\chn{M\`os\-{i}k\={e}}\xspace}
\newcommand{\pujing}{\chn{P\v{u}j\={\i}ng}\xspace}
\newcommand{\eluosi}{\chn{\'Elu\'os\=\i}\xspace}
\newcommand{\meiguo}{\chn{M\v{e}igu\'o}\xspace}
%\dangju & \dui & \shate & \jiezhe & \shizong & \yi & \an &\gandao &  \danyou \\
\newcommand{\dangju}{\chn{d\=angj\'u}\xspace}
\newcommand{\dui}{\chn{du\`\i}\xspace}
\newcommand{\shate}{\chn{Sh\=at\`e}\xspace}
\newcommand{\jiezhe}{\chn{j\`{\i}zh\v{e}}\xspace}
\newcommand{\shizong}{\chn{sh\={\i}z\=ong}\xspace}
\newcommand{\danyou}{\chn{d\=any\=ou}\xspace}
\newcommand{\gandao}{\chn{g\v{a}nd\`ao}\xspace}
\newcommand{\an}{\chn{\`an}\xspace}
\newcommand{\yi}{\chn{y\=\i}\xspace}
\newcommand{\buman}{\chn{b\`um\v{a}n}\xspace}

\newcommand{\Prob}{\ensuremath{\mathrm{P}}\xspace}

\newcommand{\lhs}{\ensuremath{\mathit{lhs}}\xspace}
\newcommand{\rhs}{\ensuremath{\mathit{rhs}}\xspace}

\newcommand{\NULL}{\ensuremath{\mathit{null}}\xspace}

\newcommand{\gap}{\ensuremath{\sqcup}\xspace}
\newcommand{\spanof}{\ensuremath{\mathit{span}}\xspace}
\newcommand{\yield}{\ensuremath{\mathit{yield}}\xspace}
\newcommand{\fs}{\ensuremath{\mathit{admset}}\xspace}

\newcommand{\closed}{\ensuremath{\mathit{closed}}\xspace}
\newcommand{\open}{\ensuremath{\mathit{open}}\xspace}
\newcommand{\hs}{\frag} %\ensuremath{\mathit{frag}}\xspace}
\newcommand{\vs}{\ensuremath{\mathit{vs}}\xspace}
\newcommand{\frag}{\ensuremath{\mathit{frag}}\xspace}
\renewcommand{\root}{\ensuremath{\mathit{root}}\xspace}
\newcommand{\leaves}{\yield} %\ensuremath{\mathit{leaves}}\xspace}
\newcommand{\exps}{\ensuremath{\mathit{front}}\xspace} % frontier
\newcommand{\newexps}{\ensuremath{\mathit{\exps'}}\xspace}

\newcommand{\PLM}{\ensuremath{\Prob_{\mathrm{lm}}}\xspace}
\newcommand{\PT}{\ensuremath{\Prob}\xspace}
\newcommand{\PLex}{\ensuremath{\Prob_{\mathrm{lex}}}\xspace}

\newcommand{\GHKM}[3]{\ensuremath{{\mbox{#1}}_{{#2}}^{{#3}}}\xspace}
\newcommand{\tspan}[1]{\scriptsize {#1}\xspace}
\newcommand{\wnode}[1]{\rnode[t]{#1}{#1}\xspace}  %% target words

\newcommand{\vtitem}[2]{\ensuremath{({\vtnt{#1}}_{#2})}\xspace}

\newcommand{\ruleset}{\ensuremath{\mathcal{R}}\xspace}

% pattern-match
\newcommand{\match}{\ensuremath{\mathit{match}}\xspace}
\newcommand{\vars}{\ensuremath{\mathit{vars}}\xspace}
\newcommand{\mlist}{\ensuremath{\mathit{mlist}}\xspace}
\newcommand{\eP}{\ensuremath{e_{\mathit{p}}}\xspace}
%\newcommand{\PLM}{\ensuremath{\Prob_{\mathrm{lm}}}\xspace}
% \newcommand{\PT}{\ensuremath{\Prob}\xspace}
% \newcommand{\PLex}{\ensuremath{\Prob_{\mathrm{lex}}}\xspace}

%NOW MOVED HERE
\newcommand{\npshift}{\hspace{1.6cm}}
\newcommand{\styleb}{linestyle=dashed}

\newcommand{\et}{\ensuremath{e^{\mathrm{t}}}}
\newcommand{\Ht}{\ensuremath{H^{\mathrm{t}}}}
\newcommand{\ep}{\ensuremath{e}\xspace}  % just for EMNLP
\newcommand{\Hp}{\ensuremath{H^{\mathrm{p}}}}
\newcommand{\Vp}{\ensuremath{V^{\mathrm{p}}}}

\def\namecite{\newcite}

\def\Fone{F$_1$\xspace}
\def\Foneperc{F$_1\%$}
\newcommand{\goto}{\ensuremath{\triangleright}\xspace}

\newcommand{\TODO}[1]{\textbf{TODO:} \textit{#1}\xspace}

\newcommand{\smallnt}[1]{\ensuremath{_{\mbox{\tiny PP}}}\xspace}

\newcommand{\shift}{\ensuremath{\mathsf{sh}}\xspace}
\newcommand{\reduce}{\ensuremath{\mathsf{re}}\xspace}
\newcommand{\lreduce}{\ensuremath{\mathsf{re_{\small \curvearrowleft}}}\xspace}
\newcommand{\rreduce}{\ensuremath{\mathsf{re_{\small \curvearrowright}}}\xspace}

\newcommand{\sep}{\ensuremath{ \circ }\xspace}

\newcommand{\sttop}{\ensuremath{s_0}\xspace}
\newcommand{\stnext}{\ensuremath{s_1}\xspace}
\newcommand{\stthird}{\ensuremath{s_2}\xspace}
\newcommand{\qhead}{\ensuremath{q_0}\xspace}
\newcommand{\qnext}{\ensuremath{q_1}\xspace}
\newcommand{\qthird}{\ensuremath{q_2}\xspace}
\newcommand{\Tag}[1]{\ensuremath{{#1}.\mathsf{t}}\xspace}
\newcommand{\Wrd}[1]{\ensuremath{{#1}.\mathsf{w}}\xspace}
\newcommand{\LC}[1]{\ensuremath{{#1}.\mathsf{lc}}\xspace}
\newcommand{\RC}[1]{\ensuremath{{#1}.\mathsf{rc}}\xspace}

% Algorithm 3 -> Pseudocode 3
\newcommand{\pseudocode}{Algorithm}
\floatname{algorithm}{\pseudocode}

\newcommand{\cont}[2]{\ensuremath{\mathit{c}({#1}, {#2})}\xspace}
\newcommand{\contR}[2]{\ensuremath{\mathit{c}_{\mbox{\tiny \sc R}}({#1}, {#2})}\xspace}
\newcommand{\contsttops}{\ensuremath{\cont{\stnext}{\sttop}}\xspace}
\newcommand{\contsttopqhead}{\ensuremath{\contR{\sttop}{\qhead}}\xspace}

\newcommand{\sspan}{\ensuremath{\mathit{sp}}\xspace}
\renewcommand{\tspan}{\ensuremath{\mathit{tp}}\xspace}
\newcommand{\aspan}{\ensuremath{\mathit{ap}}\xspace}

% vanilla non-dp shift-reduce item: (l, S, Q)
\newcommand{\olditem}[3]{\ensuremath{{#1}: \tuple{{#2}, \ {#3}}}\xspace}
\newcommand{\newitem}[4]{\ensuremath{{#1}: \tuple{{#2}, \ {#3}, \ {#4}}}\xspace}

\newcommand{\arcs}{\ensuremath{\mathbf{a}}\xspace}
\newcommand{\arcleft}[2]{\ensuremath{{#1}^\curvearrowleft{#2}}\xspace}
\newcommand{\arcright}[2]{\ensuremath{{#1}^\curvearrowright{#2}}\xspace}

% kernel feature function
\newcommand{\vecfker}{\ensuremath{\widetilde{\mathbf{f}}}\xspace}
\newcommand{\fker}[1]{\ensuremath{\vecfker({#1})}\xspace}
\newcommand{\earleyitem}[3]{\ensuremath{\tuple{{#1}, {#2}, \ {#3}}}\xspace}
\newcommand{\nodpearleyitem}[2]{\ensuremath{\tuple{{#1}, \ {#2}}}\xspace}

\iffalse
\newcommand{\eisneritem}[4]{\newitem{#1}{#2}{#3}{\!\!\!\raisebox{0.04in}{\Tree[.{\ensuremath{#4}} !\Troof{\ensuremath{{#2}...{#3}}} ]}}\xspace}
\newcommand{\specialeisneritem}[4]{\newitem{#1}{#2}{#3}{#4}\xspace}
\else
\newcommand{\eisneritem}[4]{\tuple{{#1}, {\!\!\!\raisebox{0.04in}{\Tree[.{\ensuremath{#4}} !\Troof{\ensuremath{{#2}...{#3}}} ]}}\!\!\!}\xspace}
\newcommand{\specialeisneritem}[4]{\ensuremath{{#1}: \tuple{{#4}}}\xspace}
\newcommand{\gpeisneritem}[5]{\ensuremath{{#1}:\tuple{{#4}, {\!\!\!\raisebox{0.04in}{\Tree[.{\ensuremath{#5}} !\Troof{\ensuremath{{#2}...{#3}}} ]}}\!\!\!}}\xspace}
\fi

\newcommand{\decode}{\ensuremath{\mathit{decode}}\xspace}
\newcommand{\GEN}{\ensuremath{\mathcal{Y}}\xspace}
\newcommand{\GENbar}{\ensuremath{\overline{\GEN}}\xspace}

\newcommand{\spb}[1]{\ensuremath{^{{#1}}}\xspace}
\newcommand{\spbkp}{\ensuremath{\spb{k+1}}\xspace}
\newcommand{\vecwk}{\ensuremath{\vecw\spb{k}}\xspace}
\newcommand{\vecwkp}{\ensuremath{\vecw\spb{k+1}}\xspace}
\newcommand{\vecwt}{\ensuremath{\vecw\spb{t}}\xspace}
\newcommand{\vecwtp}{\ensuremath{\vecw\spb{t+1}}\xspace}
\newcommand{\tth}{\ensuremath{t^{\mathrm{th}}}\xspace}

\newcommand{\err}{\ensuremath{\mathit{err}}\xspace}
\newcommand{\bPhi}{\ensuremath{\mathbf{\Phi}}\xspace}
\newcommand{\dPhi}{\ensuremath{\Delta\bPhi}\xspace}
\newcommand{\labels}{\ensuremath{\mathit{labels}}\xspace}
\newcommand{\onestep}{\ensuremath{\mathit{onestep}}\xspace}
\newcommand{\seqx}{\ensuremath{\overline{x}}\xspace}
\newcommand{\seqy}{\ensuremath{\overline{y}}\xspace}
\newcommand{\seqz}{\ensuremath{\overline{z}}\xspace}

\providecommand{\norm}[1]{\lVert#1\rVert}
\providecommand{\card}[1]{\lvert#1\rvert}  % cardinality |x|
\newcommand{\normtwo}[1]{\norm{#1}^2}

\newcommand{\seqzt}{\ensuremath{\overline{z}^{(t)}}\xspace}
\newcommand{\seqxt}{\ensuremath{\overline{x}^{(t)}}\xspace}
\newcommand{\seqyt}{\ensuremath{\overline{y}^{(t)}}\xspace}

\newcommand{\defeq}{\ensuremath{\stackrel{\Delta}{=}}\xspace}
\newcommand{\InlineIf}[2]{{\bf if} {#1} {\bf then} {#2}}
\newcommand{\dsep}{\ensuremath{\mathcal{D}(\vecu,\delta)}\xspace}
\newcommand{\dgsep}{\ensuremath{\mathcal{D}_g(\vecu,\delta)}\xspace}

\newcommand{\hinge}{\ensuremath{h_{\delta}}\xspace}

\newcommand{\Beam}{\ensuremath{\mathcal{B}}\xspace}

% equivalence class under ~: [[x]]_~
\newcommand{\eqvcls}[1]{\ensuremath{\llbracket {#1} \rrbracket_{\equiv}}\xspace}

\newcommand{\Cs}{\ensuremath{C_{\mathrm{s}}}}
\newcommand{\Cg}{\ensuremath{C_{\mathrm{g}}}}
\newcommand{\Cb}{\ensuremath{C_{\mathrm{b}}}}

\newcommand{\state}{\ensuremath{\mathit{state}}\xspace}
\newcommand{\action}{\ensuremath{\mathit{action}}\xspace}

\newcommand{\fkerdp}[1]{\ensuremath{\vecfker_{{\mathrm{DP}}}({#1})}\xspace}
\newcommand{\fkernodp}[1]{\ensuremath{\vecfker_{\mathrm{noDP}}({#1})}\xspace}

\newcommand{\score}{\ensuremath{\mathit{S}}\xspace}
\newcommand{\scoring}[2]{\ensuremath{\score_{#1}({#2})}\xspace}
\newcommand{\scoresh}[1]{\ensuremath{\scoring{\shift}{#1}}\xspace}
\newcommand{\scoreleft}[1]{\ensuremath{\scoring{\lreduce}{#1}}\xspace}
\newcommand{\scoreright}[1]{\ensuremath{\scoring{\rreduce}{#1}}\xspace}

\newcommand{\leftptrs}{\ensuremath{\mathcal{L}}\xspace}
\newcommand{\rightptrs}{\ensuremath{\mathcal{R}}\xspace}

\newcommand{\chart}{\ensuremath{\mathit{C}}\xspace}
\newcommand{\pq}{\ensuremath{\mathit{Q}}\xspace}

\newcommand{\bleu}{{\sc Bleu}\xspace}
\newcommand{\bleuplusone}{\ensuremath{\mathrm{Bleu}^{+1}}\xspace}
\newcommand{\mert}{{\sc Mert}\xspace}
\newcommand{\ce}{{\sc Ch-En}\xspace}
\newcommand{\se}{{\sc Sp-En}\xspace}
\newcommand{\pro}{{\sc Pro}\xspace}
\newcommand{\mira}{{\sc Mira}\xspace}
\newcommand{\hols}{{\sc Hols}\xspace}
\newcommand{\ramp}{{\sc Ramp}\xspace}
\newcommand{\hiero}{{\sc Hiero}\xspace}
\newcommand{\cdec}{{\tt cdec}\xspace}

\newcommand{\MaxForce}{{\sc MaxForce}\xspace}

\newcommand{\myinfer}[3]{\ensuremath{\infer[\mbox{\footnotesize ${#1}$}]{#2}{#3}}\xspace}
\newcommand{\pre}{\ensuremath{\mathit{pre}}\xspace}
\newcommand{\good}{\ensuremath{\mathit{good}}\xspace}
\newcommand{\bad}{\ensuremath{\mathit{bad}}\xspace}
\newcommand{\ygood}{\ensuremath{\mbox{$y$-good}}\xspace}
\newcommand{\ybad}{\ensuremath{\mbox{$y$-bad}}\xspace}

\newcommand{\cn}{\ensuremath{{\sf cn}}\xspace}
\newcommand{\en}{\ensuremath{{\sf en}}\xspace}

\newcommand{\kNN}{\ensuremath{\mbox{$k$-NN}}\xspace}

\newcommand{\Yahoo}{{\sc Yahoo}\xspace}
\newcommand{\TREC}{{\sc TREC}\xspace}
\newcommand{\DMV}{{\sc DMV}\xspace}
\newcommand{\Insurance}{{\sc Insurance}\xspace}

\def\lr{\mbox{\begin{picture}(7,10)
\put(1,0){\line(1,0){5}}
\put(1,0){\line(1,2){5}}
\put(6,0){\line(0,1){10}}
\end{picture}
}}

\def\ll{\mbox{\begin{picture}(7,10)
\put(1,0){\line(1,0){5}}
\put(6,0){\line(-1,2){5}}
\put(1,0){\line(0,1){10}}
\end{picture}
}}

\def\llbig{\mbox{\begin{picture}(25,10)
\put(1,0){\line(1,0){30}}
\put(31,0){\line(-3,1){30}}
\put(1,0){\line(0,1){10}}
\end{picture}
}}

\def\lrbig{\mbox{\begin{picture}(25,10)
\put(1,0){\line(1,0){30}}
\put(1,0){\line(3,1){30}}
\put(31,0){\line(0,1){10}}
\end{picture}
}}
\newcommand{\doubleitem}[2]{\ensuremath{\stackrel{\!\!\!\!\!{#1}}{\ll} \stackrel{\ {#2}}{\lr}}\xspace}

\newcommand{\doubleitemleft}[2]{\ensuremath{\stackrel{\hspace{-0.9cm}{#1}}{\llbig} \quad \stackrel{\ {#2}}{\lr}}\xspace}

\newcommand{\doubleitemright}[2]{\ensuremath{\stackrel{\!\!\!\!\!{#1}}{\ll} \; \stackrel{\hspace{1.1cm} {#2}}{\lrbig}}\xspace}

\newcommand{\pytorch}{PyTorch\xspace}
\newcommand{\rnnsearch}{RNNsearch\xspace}

\newcommand{\best}{\ensuremath{\mathit{best}}\xspace}
\newcommand{\bestuptoi}{\ensuremath{\mathit{best}_{\leq i}}\xspace}
\newcommand{\completed}{\ensuremath{\mathit{comp}}\xspace}
\newcommand{\eos}{\mbox{\scriptsize \texttt{<eos>}}\xspace}
\newcommand{\bp}{\ensuremath{\mathit{bp}}\xspace}
\newcommand{\BLEU}{\ensuremath{\mathrm{BLEU}}\xspace}
\newcommand{\PWR}{\ensuremath{\mathrm{PWR}}\xspace}
\newcommand{\AdaR}{\ensuremath{\mathrm{AdaR}}\xspace}
\newcommand{\gnmt}{\ensuremath{\mathrm{GNMT}}\xspace}
\newcommand{\lengthnorm}{\ensuremath{\mathrm{length\_norm}}\xspace}
\newcommand{\word}{\ensuremath{\mathrm{word}}\xspace}
\newcommand{\len}{\ensuremath{\mathit{len}}\xspace}
\newcommand{\gratio}{\ensuremath{\mathit{gr}}\xspace}
\newcommand{\lratio}{\ensuremath{\mathit{lr}}\xspace}
\newcommand{\pred}{\ensuremath{\mathit{pred}}\xspace}

\newcommand{\spinyin}[1]{{\small\em {#1}}}
\newcommand{\bajisitan}{B\=aj\={\i}s\={\i}t\v{a}n}
\newcommand{\yindu}{Y\`{\i}nd\`u}

\newcommand{\deen}{{{de}$\leftrightarrow${en}}\xspace}
\newcommand{\zhen}{{{zh}$\leftrightarrow${en}}\xspace}
\newcommand{\detoen}{{{de}$\goesto${en}}\xspace}
\newcommand{\entode}{{{en}$\goesto${de}}\xspace}
\newcommand{\zhtoen}{{{zh}$\goesto${en}}\xspace}
\newcommand{\entozh}{{{en}$\goesto${zh}}\xspace}

\newcommand{\uarrow}{\ensuremath{\uparrow}\xspace}
\newcommand{\darrow}{\ensuremath{\downarrow}\xspace}
\newcommand{\larrow}{\ensuremath{\leftarrow}\xspace}
\newcommand{\rarrow}{\ensuremath{\rightarrow}\xspace}

\newcommand{\floor}[1]{\lfloor #1 \rfloor}
\newcommand{\gwaitk}{\ensuremath{{g_\text{wait-$k$}}}\xspace}
\newcommand{\gcatchup}{\ensuremath{{g_\text{wait-$k$, $c$}}}\xspace}

\newcommand{\CW}{\ensuremath{\mathrm{CW}}\xspace}
\newcommand{\AP}{\ensuremath{\mathrm{AP}}\xspace}
\newcommand{\AL}{\ensuremath{\mathrm{AL}}\xspace}
\newcommand{\NE}{\ensuremath{\mathrm{NE}}\xspace}

\newcommand{\thetafull}{\ensuremath{\bm{\theta}_\text{full}}\xspace}
\newcommand{\thetafullmt}{\ensuremath{\thetafull^\text{MT}}\xspace}
\newcommand{\thetafullmthat}{\ensuremath{\hat{\bm{\theta}}_\text{full}^\text{MT}}\xspace}

\newcommand{\thetafullst}{\ensuremath{\thetafull^\text{ST}}\xspace}
\newcommand{\thetafullsthat}{\ensuremath{\hat{\bm{\theta}}_\text{full}^\text{ST}}\xspace}
\newcommand{\thetafullasrhat}{\ensuremath{\hat{\bm{\theta}}_\text{full}^\text{ASR}}\xspace}
\newcommand{\thetafullasr}{\ensuremath{{\bm{\theta}}_\text{full}^\text{ASR}}\xspace}

\newcommand{\thetawk}{\ensuremath{\bm{\theta}^{\text{wait-$k$}}}}
\newcommand{\hatpfull}{\ensuremath{\hat{p}_{\text{full}}}}
\newcommand{\pwk}{\ensuremath{p_{\text{wait-$k$}}}}

\newcommand{\nextbeam}{\ensuremath{\mathit{next}\xspace}}
\newcommand{\tabincell}[2]{\begin{tabular}{@{}#1@{}}#2\end{tabular}}
\newcommand{\jc}[1]{{\color{blue}{[{\bf junkun}: #1]}}}

% \newcommand{\pseudocode}{Algorithm}
% \floatname{algorithm}{\pseudocode}
\newcommand{\bmtheta}{\ensuremath{\bm{\theta}}}
\newcommand{\waitk}{\ensuremath{\text{wait-$k$}}\xspace}
\newcommand{\HR}{\ensuremath{\mathit{HR}}\xspace}
\newcommand{\pfull}{\ensuremath{p_{\text{full}}}}
\newcommand{\psim}{\ensuremath{p_{\text{simul}}}}

\newcommand{\dsxy}{D_{\vecs,\vecx,\vecy}\xspace}
\newcommand{\dsy}{D_{\vecs,\vecy}\xspace}
\newcommand{\dsx}{D_{\vecs,\vecx}\xspace}
\newcommand{\dxy}{D_{\vecx,\vecy}\xspace}
\newcommand{\ds}{D_{\vecs}\xspace}
\newcommand{\dx}{D_{\vecx}\xspace}
\newcommand{\dy}{D_{\vecy}\xspace}
\newcommand{\rw}{\ensuremath{\textit{RW}}\xspace}
\newcommand{\cp}{\ensuremath{\textit{CP}}\xspace}
\newcommand{\cs}{\ensuremath{\mathit{u}}\xspace}
\newcommand{\ycandidate}{\ensuremath{\hat{\vecy}}\xspace}
\newcommand{\sameabove}{\raisebox{0.5ex}{\texttildelow}\xspace}
\section{Introduction}
\label{sec:intro}

Simultaneous Speech-to-Text translation (ST) incrementally translates speech in a source language speech into text in a target language, 
and has found wide-ranging applications in numerous crosslingual communication scenarios 
such as international travel and multinational conferences. 

Unlike the full-sentence translation, 
which translates upon the cessation of speech segments,
and provides a complete translation for an entire segment.
Simultaneous ST requires the generation of intermediate translations as the speech continues.
These partial results are of critical importance, 
enabling the audience to keep pace with the content of the speaker's discourse in real time, 
fostering immediate comprehension and engagement.

In recent years, the end-to-end (E2E) approach has surpassed conventional cascaded methods
in terms of performance \cite{ren2020simulspeech,chen2021direct}. 
Notably, the implementation of Transducer models for the adaptive simultaneous translation for streaming speech 
has significantly enhanced translation quality \cite{xue2022large,tang2023hybrid}. 
Despite these advancements, the stability issue remains unaddressed in this task.

\begin{figure}[t]\centering
    \resizebox{\linewidth}{!}{
    %\centering
    \setlength{\tabcolsep}{1.5pt}
    \renewcommand{\arraystretch}{1.}
    \begin{tabu}{c | l l l l l l l l l l l}
    
    \toprule
    \rowfont{\small}
    & \textit{m\v{e}igu\'o} & \textit{de} & \textit{zh\=ong} & \textit{x\=i} & \textit{b\`u} & \textit{y\v{o}u} & \textit{h\v{e}ndu\=o} & \textit{g\=ao} & \textit{sh\=an} & &\\[-0.1cm]
    \tabincell{c}{Source\\ Transcription} & 美国 & 的 & 中 &西 & 部 & 有 & 很多 & 高 & 山   \\[-0.2cm]
     \rowfont{\small\it}
     & USA & 's & central & west & area & have & many & big & mountain\!\!\!\!\! & \\
    \midrule \midrule
    Translation-Ref & \multicolumn{10}{l}{\textbf{there \ are \ many \ big \ mountains \ in \ west \ central \ US}}\\
    \midrule \midrule
    \multirowcell{6}{(a) \\ E2E Streaming \\ Translation}& \multicolumn{10}{c}{{\small \it (audio and segment start)}} \\
    & \multicolumn{1}{c}{{\small\it [$t_1$]}} & \multicolumn{9}{l}{\textbf{\color{blue}{American}}}  \\ 
    & \multicolumn{1}{c}{{\small\it [$t_2$]}} & \multicolumn{9}{l}{\textbf{\color{blue}{West \ central \ US}}}  \\ 
    & \multicolumn{1}{c}{{\small\it [$t_3$]}} & \multicolumn{9}{l}{\textbf{West \ central \ US \ \color{blue}{has \ many}}}  \\ 
    & \multicolumn{1}{c}{{\small\it [$t_3$]}} & \multicolumn{9}{l}{\textbf{\color{blue}{there \ are \ many \ big \ mountains \ in \ west \ central \ US}}} \\
    & \multicolumn{10}{c}{{\small \it (audio and segment end)}} \\
    \midrule
    \multirowcell{6}{(b) \\ Revision-Free \\ Decoding}& \multicolumn{10}{c}{{\small \it (audio and segment start)}} \\
    & \multicolumn{1}{c}{{\small\it [$t_1$]}} & \multicolumn{9}{l}{\textbf{\color{blue}{American}}}  \\ 
    & \multicolumn{1}{c}{{\small\it [$t_2$]}} & \multicolumn{9}{l}{\textbf{American \ \color{blue}{midwest}}}  \\ 
    & \multicolumn{1}{c}{{\small\it [$t_3$]}} & \multicolumn{9}{l}{\textbf{American \ midwest \ \color{blue}{has \ many}}}  \\ 
    & \multicolumn{1}{c}{{\small\it [$t_3$]}} & \multicolumn{9}{l}{\textbf{American \ midwest \ has \ many \ \color{blue}{big \ mountains}}} \\
    & \multicolumn{10}{c}{{\small \it (audio and segment end)}} \\
    \bottomrule
    \end{tabu}
    } % resizebox
    \caption{
     A decoding example of E2E simultaneous ST. 
     The provided source transcription represents the content of the 
     source speech in Chinese, 
     with the corresponding gloss also displayed.
     Text in \textcolor{blue}{\textbf{blue}} denote newly generated translation.
     The standard approach showcases the possibility for intermediate translation to flicker with continuous speech input. 
     In contrast, our proposed revision-free decoding method strives to maintain the intermediate translation unrevised. 
     % for wait-3 policy.
    \label{fig:idea}
    }
    
    %% \\vspace{-0.5cm}
    \end{figure}

As the speech continues, 
a simultaneous ST system does not inherently guarantee to append new words to the previous partial result. 
As shown in Figure \ref{fig:idea}(a), words previously displayed can be removed or altered.
This instability in the partial results can lead to frequent alterations on screen, 
causing the results to flicker. 
While permitting revisions has the potential to improve translation quality, 
this flickering creates an unfavorable user experience
and can be distracting \cite{baumann2009assessing,stolte2021automatic}.
It causes discomfort among audience members, 
who might consequently lose track of the content.
Given the reordering nature between different languages \cite{birch+:2009,braune+:2012}, 
the experience with flickering is substantially worse 
than that of Automatic Speech Recognition (ASR) tasks, 
which maintain a monotonic alignment.

Furthermore, simultaneous ST is usually succeeded by an incremental Text-To-Speech (TTS) system 
that synthesizes the text into speech in the target language \cite{zheng2020fluent}. 
Since the synthesized speech display cannot be retroactively altered, 
flickering poses significant challenges.

In contrast to the ASR task, 
which invariably aims for a single optimal outcome as the desired recognition result. 
The ST can accommodate multiple valid translations for a single input. 
For instance, as shown in Fig.~\ref{fig:idea}, 
both ``\textit{American midwest}'' and ``\textit{West central US}'' 
can serve as suitable translations for ``\textit{美国的中西部}''. 
Furthermore, 
the flexibility of languages allows for multiple equivalent expressions \cite{koehn2009statistical}.
Therefore, not all revisions are necessary. 
To maintain the stability of partial results, 
a continuous translation strategy can be implemented, 
which builds upon the previously generated prefix.
As illustrated in Fig.~\ref{fig:idea}(b), 
a revision-free decoding approach, 
which refrains from modifying generated partial results, 
can also yield translations of considerable quality.

While greedy decoding is widely utilized in simultaneous translation to guarantee stability \cite{ma+:2019}, 
it often compromises the overall translation quality \cite{zheng2019speculative}. 
In this work, 
we propose two strategies aimed at mitigating the flickering issues observed in simultaneous ST.
We identify the root cause of flickering in the ranking process of beam search decoding. 
It presents a potential avenue for reducing the frequency of 
intermediate translation commitments and thereby preventing unnecessary revisions. 
Furthermore,
to fundamentally address the flickering issue with beam search,
we introduce a novel revision-controllable approach that actively manages revisions during the translation decoding process. 
Our primary idea is to modify the beam search pruning process through the introduction of an allowed revision window. 
It can filter out candidates that may induce extensive revisions. 
Our method can completely prevent flickering during decoding with only a minor reduction in translation quality. 

By adjusting the allowed revision window, 
our proposed method is capable of achieving performance comparable to existing methods in terms of translation quality, 
while significantly enhancing both latency and stability.
This dual strategy presents a promising solution for enhancing the performance and user experience of simultaneous ST.

\section{Preliminary}
\label{sec:prelim}

We first briefly review end-to-end simultaneous speech translation
to set up the notations.

\subsection{End-to-End Speech Translation}
Regardless of the specific model architecture employed, 
the novel paradigm of E2E ST delineates an objective: 
to transform a speech feature sequence \vecx $= (x_1,x_2,...,x_T) $ into a series of text tokens \vecy in a different language,
where each $x_i$ represents the frame-level feature with a certain duration. 

The conventional cascaded system uses an ASR model to convert speech to text in the source language,
which is subsequently fed into a machine translation (MT) model to get the translation in the target language.
Unlike this method,
the E2E ST model incorporates a singular, 
integrated model which does not need an intermediate recognized result,
thereby alleviating the issue of error propagation \cite{sperber2019attention}.
The model can be formalized as:
\begin{align*}
  \pfull(\vecy \mid \vecx;\bmtheta) = \prod\limits_{t=1}^{|\vecy|}p(y_t \mid \vecx, \vecy_{<t};\bmtheta) 
  % \label{eq:fullsentence_decode}
\end{align*}

\subsection{Simultaneous End-to-End Speech Translation with Transducer Model}
\label{subsec:simul_ST}
Simultaneous E2E ST system translates concurrently with continuous source speech.
Formally, the prediction of \vecy can be defined as,
\begin{align*}
  \psim(\vecy \mid \vecx;\bmtheta) = \prod\limits_{t=1}^{|\vecy|}p(y_t \mid \vecx_{<\tau(t)}, \vecy_{<t};\bmtheta) 
  % \label{eq:fullsentence_decode}
\end{align*}
where $\tau(t)$ denotes the timestamp to decode target token $y_t$.

\begin{figure}[t]
    \centering
    \includegraphics[width=1\linewidth]{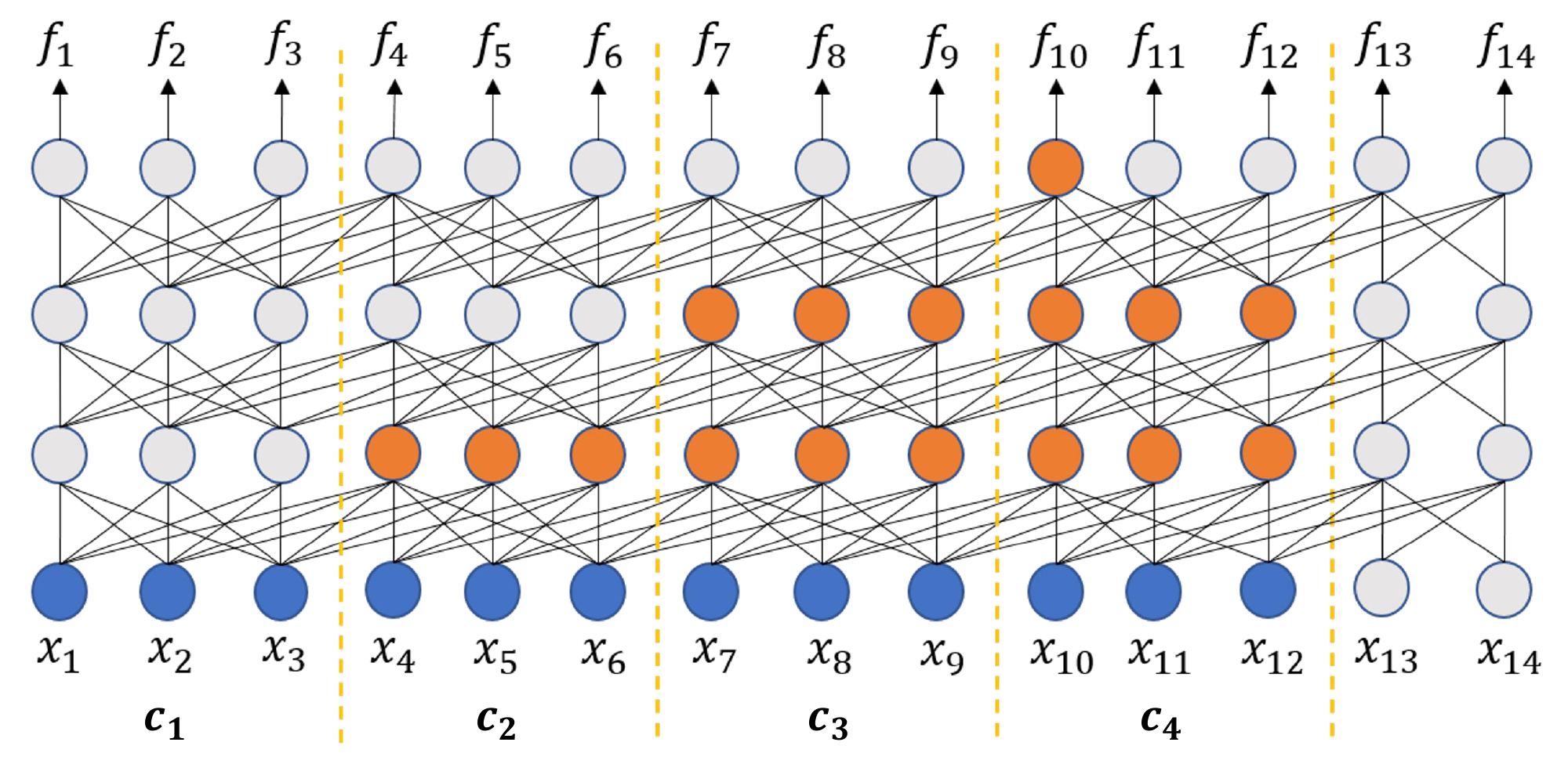}
    \caption{TT encoding at position $f_{10}$, utilizing an attention mask. 
    The process is characterized by a specified chunk size of $3$ and   
     the number of left chunk $1$.}
    %Note our translation model does not use any ASR output as input.}
    \label{fig:TT}
    \end{figure}

Recently, the neural transducer model \cite{graves2012sequence} presents a compelling fit for simultaneous E2E ST \cite{xue2022large, tang2023hybrid}.
Its design considers all potential alignments between speech and text throughout the training process,
and it shows the capacity to adaptively translate speech into text in a streaming manner \cite{papi2023token}.
In \cite{xue2022large},
the authors proposed to leverage the Transformer-Transducer (T-T) \cite{yeh2019transformer, zhang2020transformer} for simultaneous ST. To realize the low-latency high-accuracy streaming T-T, 
they use the speech encoder design in \cite{chen2021developing}, shown in Fig.~\ref{fig:TT}. 

To uphold low latency and minimize computational costs,
the input speech frames \vecx are segmented and sequentially fed into the encoder in distinctive chunks \vecc.
Each chunk $\vecc_i$ comprises several speech frames,
the quantity of which aligns with the chunk size \cs.
The incorporation of the attention masks \cite{chen2021developing} facilitates processing in a chunk-wise streaming fashion. 
Every frame has a predetermined number of visible left chunks, 
and the size of the left reception field grows proportionally with the number of layers. 
This allows the model to leverage extensive historical information for improved performance 
while significantly reducing computational requirements compared to 
models that consider the entire history at each layer. 
Within a chunk, 
all frames can observe one another, 
but they cannot access frames in subsequent chunks.

\subsection{Decoding with Beam Search}
\label{subsec:beam_search}
In order to achieve a fluent translation, 
the application of beam search is crucial during the decoding process.
We denote $B_i$ to be the beam at time step $i$, which is an ordered list with a beam
size of $b$, and it expands to the next beam $B_{i+1}$ with the same size:
\begin{align*}
%\vspace{-0.2cm}
B_0 =& [\tuple{\startsym, \psim(\startsym \mid x_1;\bm{\theta})}] \\
B_i =& \textstyle\toptop ^b  ( \nextbeam (B_{i-1}, i) )\\
\nextbeam (B, i)  = & 
     \{\tuple{\vecy \circ y_i, \psim(y_i \mid \vecx_{\leq \tau(i)}, \vecy; \bm{\theta})} \mid\\
     &  \;\; \tuple{\vecy, p} \in B, y_i \in V \} \\
  %    \vspace{-0.3cm}
\ycandidate_i &= \textstyle\toptop ^1  ( \nextbeam (B_{i-1}, i) )[0]
\end{align*}
The best hypothesis $\ycandidate_i$ is usually used as the intermediate generated result \cite{yu2016automatic}.
Intermediate result revision happens when the top candidate $\ycandidate_i$ in $B_i$ is neither identical to nor a prefix of the top candidate $\ycandidate_{i+1}$ in $B_{i+1}$.
\section{Methods}
\label{sec:method}

As the simultaneous ST model processes continuous speech, 
it generates intermediate translation results. 
In the context of online decoding utilizing beam search, 
the intermediate translation is typically represented by the best candidate \ycandidate in the beam. 
However, as discussed in Sec.~\ref{subsec:beam_search}, 
these best candidates may not always serve as the prefix for the subsequent translation. This discrepancy arises due to the inherent reranking property of the beam search algorithm, which can modify the candidate order based on their evolving scores as the decoding process advances.

\subsection{Chunk Preservation}
In conventional methods for text-based simultaneous MT, 
such as the widely-used \waitk method \cite{ma+:2019}, 
the model commits a single target token each time it receives a new source token. 
However, this approach may not be the most efficient for Speech-to-Text translation, 
where the input granularity is at the frame level. 
Individually, frames often lack sufficient semantic information, 
and their length generally exceeds that of the target text sequence.

Adopting the chunk-based model, as detailed in the Sec.~\ref{subsec:simul_ST}, 
allows the input to be processed chunk-by-chunk. 
In this approach, it is not necessary to commit results at the frame level.
Committing at the chunk level is more logical given the characteristics of ST. 
This modification offers two primary advantages:
\begin{itemize}
    \item It reduces the frequency of commitments, thereby preventing unnecessary revisions within each chunk.
    \item It streamlines the process for committing outputs, saving computational efforts and reducing system communication overhead.
\end{itemize}

Therefore, 
we proposed a method called Chunk Preservation (\cp).
For instance, as depicted in Fig.~\ref{fig:TT}, 
the proposed method refrains from committing intermediate translations for individual frames such as $f_{10}$. 
Instead, it only commits the best candidates as the intermediate translation at the end frame of each chunk, 
such as $f_{12}$ in the example. 
It can be formalized as,
\begin{align*}
\mathbf{commit}(\ycandidate_i, \cs) = 
\begin{cases} 
\text{True} & \text{if } i \bmod \cs = 0 \\
\text{False} & \text{otherwise}.
\end{cases}
\end{align*}
By committing results at the chunk level rather than the frame level, 
the translation process aligns more closely with the natural input processing pattern of ST, 
greatly enhancing the decoding stability.

It is important to note that while this method modifies the commitment approach, 
it leaves the translation quality unaffected. 
The translation accuracy remains consistent with that achieved through standard frame-level commitment.

\subsection{Revision Window Control}
\label{subsec:rw}
However, chunk preservation cannot fundamentally address the issue of revision.
Given that revisions originate from the ranking process in beam search, 
we can maintain a beam where the revisions applied to subsequent translations from all candidates are kept within a specified limit.
Thus, we design a revision-controllable decoding method, 
rooted in the beam search process, 
in which we incorporate a Revision Window (\rw).
This value regulates the maximum number of tokens that can be revised in subsequent decoding. 
Specifically, every time we commit the intermediate translation (at the end of each chunk), 
we prune the beam with this revision window control strategy in effect.

The proposed beam pruning method can be described as,
\begin{align*}
\mathbf{accept}(\vecx, \vecy, \rw) = &
\begin{cases}
\text{True}, & \text{if } \vecx_{:|\vecy|-\rw} = \vecy_{:|\vecy|-\rw} \\
\text{False}, & \text{otherwise}
\end{cases}\\
B_i =& \textstyle\toptop ^{b*}  ( \nextbeam (B_{i-1}, i) ) \\
\quad \forall \tuple{\vecy, p}  \in \nextbeam& (B_{i-1}, i), \  \mathbf{accept}(\vecy, \ycandidate_i, \rw).
\end{align*}

In essence, 
all the surviving candidates within the beam must maintain an identical prefix, 
with the length being subject to the provided revision window.
Otherwise, they will be pruned, regardless of their scores. 
\rw denotes how many tokens at the end of the intermediate translation
are permitted to be revised in the beam search process due to the progress and reranking stemming from the ongoing translation.
An extreme case occurs when $\rw=0$. 
In this case, all candidates employ the best candidates as the prefix
\footnote{it is possible that fewer than $b$ candidates survive.},
ensuring that subsequent translations will not revise the previous intermediate translation.
In scenarios that do not require strict controls on revision,
\rw can be adjusted to strike a balance between translation quality and stability.

\section{Experiments}
\label{sec:exps}

\subsection{Data and Model}
\begin{table}[ht]
% \resizebox{\linewidth}{!}{
% \begin{tabular}{c|cccccc}
% \toprule
% Language & DE  & ES  & IT  & FR  & NL  & ZH    & EN   \\
% \midrule
% Hours    & 29k & 33k & 32k & 28k & 10k & 129k  & 230k \\
% \bottomrule
% \end{tabular}

% } % resize box
\centering
\resizebox{0.9\linewidth}{!}{
\begin{tabular}{l|cccccc}
\toprule
Language & DE  & ES  & IT  & FR  & NL  & ZH   \\
\midrule
Hours    & 15k & 17k & 16k & 14k & 5k & 65k  \\
\bottomrule
\end{tabular}
} % resize box

\caption{Statistics of training speech corpora for each source language.}
\label{tab:train_data}
\end{table}

\begin{table*}[t]
\resizebox{\linewidth}{!}{
\setlength{\tabcolsep}{3.pt}
\begin{tabular}{lccc|ccc|ccc|ccc|ccc|ccc}
\toprule
  & \multicolumn{3}{c|}{DE \goesto EN} & \multicolumn{3}{c|}{ES \goesto EN} & \multicolumn{3}{c|}{IT \goesto EN} & \multicolumn{3}{c|}{FR \goesto EN} & \multicolumn{3}{c|}{NL \goesto EN} & \multicolumn{3}{c}{ZH \goesto EN} \\
\cmidrule{2-19}
  & \BLEU\uarrow    & \AL\darrow     & \NE\darrow    & \BLEU\uarrow     & \AL\darrow    & \NE\darrow    & \BLEU\uarrow     & \AL\darrow     & \NE\darrow    & \BLEU\uarrow     & \AL\darrow    & \NE\darrow   & \BLEU\uarrow     & \AL\darrow    & \NE\darrow    & \BLEU\uarrow     & \AL\darrow     & \NE\darrow    \\
\midrule
$b=1$ & 19.55 & 1317 & \underline{0.00} & 18.96 & 1239 & \underline{0.00} & 17.94 & 1270 & \underline{0.00} & 19.04 & 1112 & \underline{0.00} & 21.17 & 1183 & \underline{0.00} & 2.02 & 3140 & \underline{0.00} \\
$b=7$ & 26.28 & 1057 & 1.49 & 26.68 & 1054 & 1.74 & 26.50 & 1052 & 1.59 & 26.30 & 1061 & 1.60 & 28.28 & 967 & 1.37 & 10.97 & 1418 & 2.56 \\
\hspace{1em} + \cp & \sameabove & 1110 & 1.00 & \sameabove  & 1105 & 1.15 & \sameabove  & 1106 & 1.08 & \sameabove  & 1035 & 1.07 & \sameabove  & 1045 & 0.92 & \sameabove  & 1122 & 2.08 \\
\hspace{2em} + \rw= 0 & 25.13 & 689 & \underline{0.00} & 24.28 & 549 & \underline{0.00} & 25.18 & 648 & \underline{0.00} & 24.99 & 591 & \underline{0.00} & 27.11 & 756 & \underline{0.00} & 10.32 & 748 & \underline{0.00} \\
\hspace{2em} + \rw= 3 & 26.33 & 800 & 0.11 & 26.61 & 730 & 0.11 & 26.55 & 768 & 0.11 & 26.41 & 707 & 0.11 & 28.27 & 842 & 0.12 & 11.07 & 922 & 0.12 \\
\bottomrule
\end{tabular}
} % resize box
\caption{Performance metrics on the CoVoST2 test set across various translation directions.
\sameabove denotes that the value is the same as the one in the row above.
Chunk preservation does not change the decoding results,
it yields the same \BLEU score as the standard frame-level beam search decoding.
}
\label{table:eval_res}
\end{table*}

To demonstrate the effectiveness of our proposed method,
we conducted experiments on multiple translation directions.

Following \cite{xue2022weakly},
we trained a multilingual ST model that effectively translates speech from various languages into English. 
In the experiment,
we used a collection of anonymized internal speech corpora intended for ASR.
The specific languages covered, 
along with their corresponding durations of training data, 
are shown in Table \ref{tab:train_data}.
The training translation references were annotated with a MT service.
%In order to enhance its performance, 
%we empirically incorporated a moderate quantity of English ASR data into the joint training process. 
The model is constructed with a direct translation framework,
allowing it to seamlessly process speech in a set of languages,
specifically German (DE), Spanish (ES), Italian (IT), French (FR), Dutch (NL), and Chinese (ZH),
into English (EN) without necessitating any language-specific configuration.

We adopt the Transformer-Transducer as the foundational architecture for our model,
with a chunk size of $4$ and masked attention, as detailed in Sec.~\ref{sec:prelim},
to enable the ability to process streaming input.
We set the chunk size $\cs=4$ (i.e., 4 frames per chunk and the frame span is 40ms) and set the left chunk value to $18$. 
More specifically,
the encoder consists of a Transformer architecture with $18$ layers with $2048$ hidden size,
each having $8$ attention heads with an attention dimension of $256$.
For the prediction network,
we use a $2$-layer stacked LSTM \cite{hochreiter1997long},
with a  hidden size of $1024$,
thus allowing for efficient sequence prediction.
We set the embedding size to $320$.
The model is trained with AdamW optimizer \cite{loshchilov2018decoupled}
for $1.6$ million steps.

The efficacy of our proposed method is evaluated on the CoVoST2 \cite{wang2021covost} X$\rightarrow$En translation set, 
with individual assessments conducted for each respective language pair. 
Our evaluation metric encompasses three core dimensions: translation quality, latency, and stability.

For translation quality,
we report the case-sensitive detokenized \BLEU using \texttt{sacreBLEU}\footnote{\url{https://github.com/mjpost/sacreBLEU}} \cite{post-2018-call}.
In terms of latency, we employ Average Lagging (\AL) \cite{ma+:2019} in milliseconds.
This crucial measure enables us to understand the real-time applicability of our method in practical scenarios.
Given that the \AL metric is traditionally intended for decoding processes wherein intermediate results remain unrevised, 
and typically employed for greedy decoding methodologies, 
we adjust its application for our research context. 
In this study, 
we conduct an offline evaluation of \AL by analyzing the timestamps corresponding to each decoded token. 
Specifically, we identify the moment when a decoded token is finalized and subsequently remains unchanged for the rest of the decoding process. 
We leverage the index of the frame in which the intermediate result is committed to measure the latency (\AL). 
This differs from using the end frame of each chunk, which represents real non-computation aware latency.
Our chosen method is adopted to effectively account for the latency introduced by computational processes and display time.
Finally, for assessing the stability of our method,
we employ the metric of Normalize Erasure (\NE) \cite{arivazhagan2020re}. 
This metric quantifies the number of partial target tokens that are erased relative to each final target token.
\begin{figure}[t]
    \centering
    \begin{subfigure}{\linewidth}
        \centering
        \includegraphics[width=.95\linewidth]{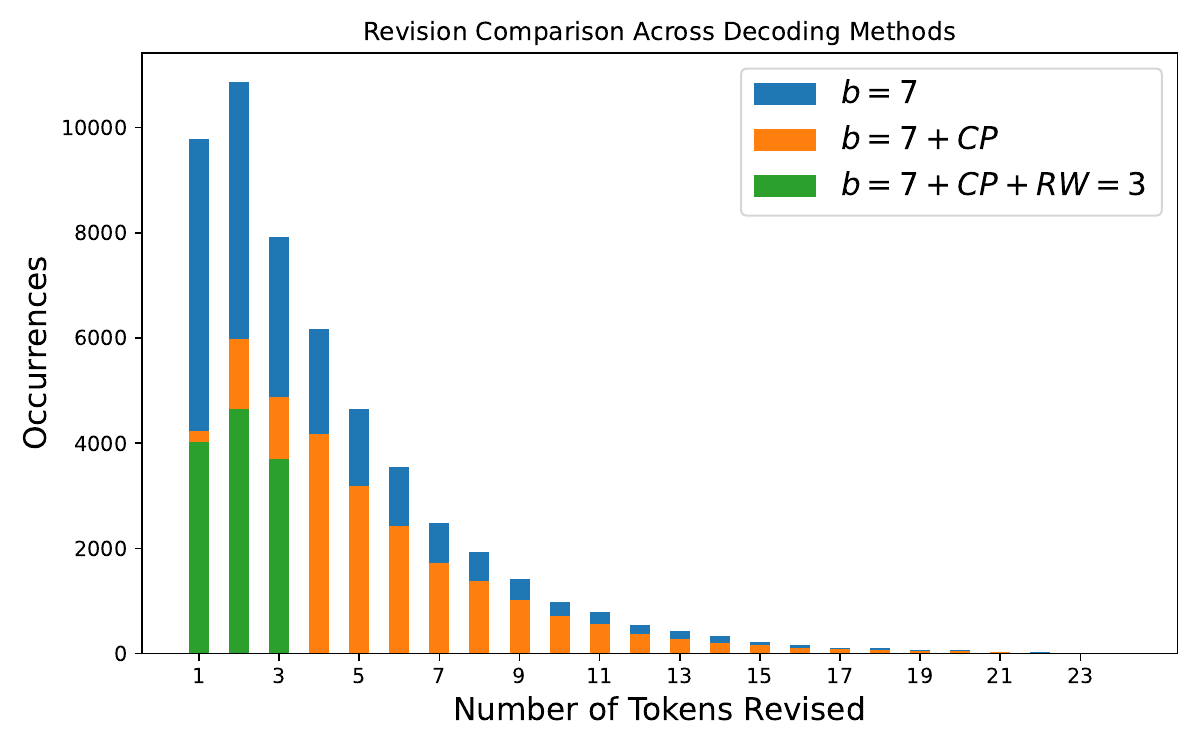}
        \caption{DE$\goesto$EN}
        \label{fig:deen}
    \end{subfigure}
    \hfill
    \begin{subfigure}{\linewidth}
        \centering
        \includegraphics[width=.95\linewidth]{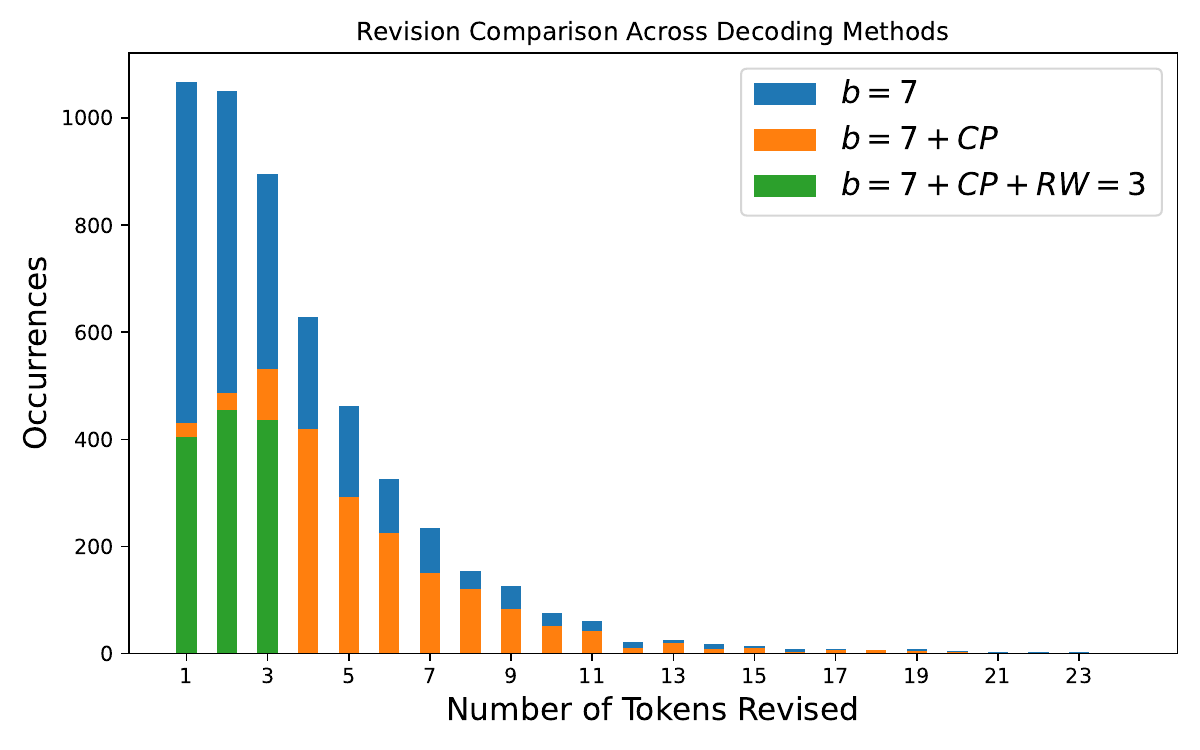}
        \caption{NL$\goesto$EN}
        \label{fig:iten}
    \end{subfigure}
    \caption{Revision count comparison across three decoding methods. This bar chart presents the comparison of decoding steps (y-axis)
    of the number of tokens been revised (x-axis) with three different decoding methods.
    % For each method, the number of occurrences for a certain number of revisions is represented on the y-axis.
    It allows for a direct comparison of how often specific numbers of revisions occur in each method.}
    \label{fig:revision_analy}
\end{figure}

\subsection{Evaluation}

Throughout the experimentation with our proposed method, 
we consistently employ a \textit{beam} size $b=7$. 
It is noteworthy that conventional beam search algorithms tend to favor shorter hypotheses, 
a characteristic often encountered in machine translation scenarios. 
To counterbalance this inherent bias towards brevity,
we incorporate a \textit{Word Reward} \cite{he+:2016} parameter set to 1 in beam search where revision window control is not employed.

The evaluation results are shown in Table \ref{table:eval_res}.
It is evident that utilizing a beam size $b=1$ ensures no revision in decoding.
However,
it markedly compromises the performance.
This approach typically generates hypotheses shorter than expected, 
primarily because shorter hypotheses tend to have better scores.
The availability of only a single candidate as the prefix throughout the entire decoding process
hinders the growth of the sequence in subsequent decoding,
given that transducer decoding follows a frame-level synchronized style,
As a result, 
it incurred a severe brevity penalty, 
leading to both poor quality and increased latency.
This underscores the critical role of beam search in decoding with transducer models,
unlike in the case of sequence-to-sequence models.
The use of chunk preservation in our model effectively mitigates flickering and improves stability, 
as indicated by the \NE scores.
Despite this minor latency increase, 
the gains in stability make this a beneficial trade-off.

The introduction of a controlled revision window plays a pivotal role in our method.
In contrast to pruning the beam to a size of 1,
our revision-controllable method is capable of maintaining multiple candidates in the beam
 as long as they do not introduce flickering beyond the given revision window in subsequent decoding.
The method can fundamentally prevent unnecessary revisions and achieves enhanced latency 
(as reflected with \rw $=0$).
Moreover, our model exhibits significant flexibility, facilitated by the adjustability of the revision window.
With \rw $=3$, 
the model is able to achieve translation quality comparable to that of beam search without revision window control.
And it still leads to significant enhancements in both latency and stability.
Especially, the enhanced stability, indicated by the lower \NE scores,
ensures the model's output consistency, 
reinforcing the reliability of the translations produced.
This clear improvement in both latency and stability, 
without compromising the translation quality, 
underlines the effectiveness of our proposed method.

\subsection{Analysis on Revision}

While our proposed methods yield a notable improvement,
to assess the stability of decoding at a more granular level, 
we performed an analysis of the frequency count for each specified number of tokens revised during the decoding process.

As demonstrated in Figure \ref{fig:revision_analy}, 
the chunk preservation method significantly mitigates the flickering issue
for both DE$\goesto$EN and NL$\goesto$EN translation, 
leading to fewer revisions during decoding and thereby enhancing NE. 
Despite these improvements, 
chunk preservation is unable to prevent extreme cases of long-range revisions 
where a large number of previously generated tokens are revised, 
drastically undermining stability.

In contrast, 
our proposed revision-controllable method effectively counteracts this problem. 
By implementing an allowable revision window (\rw), 
we establish an upper limit to the number of tokens that can be revised. 
This is accomplished via a novel pruning method within the beam search process (as detailed in Sec.~\ref{subsec:rw}), 
significantly bolstering the stability of the decoding process.

\section{Related Works}
The conventional approach to preventing flickering during the decoding process involves the use of greedy decoding, 
a technique frequently employed in text-based simultaneous MT \cite{ma+:2019,chen2021improving}. 
However, for transducer-based streaming decoding, 
this method proves unsatisfactory due to its inherent limitations on output quality.
For re-translation-style simultaneous translation models, 
a biased beam search approach has been utilized to enforce the decoding of previously generated text as a prefix \cite{arivazhagan2020re}. 

The most relevant study to our work is \cite{bruguier2023flickering}, 
the authors propose an altered approach to hypothesis selection within the beam during partial generation. 
Instead of always selecting the highest-ranked hypothesis from the beam,
they introduce a method to select the partial result,
sticking a balance between flickering, quality, and latency.
This method is employed in steaming ASR not simultaneous ST, 
which contains long-distance reorderings.
And it does not modify the beam search process.

\section{Conclusion}

In this work,
we presented a simple and effective method to 
address the issue of decoding stability in E2E simultaneous ST,
particularly, the flickering of partial translation results.
We propose two methods for reducing flickering. 
First, we introduce a straightforward technique, called chunk preservation, 
which significantly reduces flickering while maintaining the translation quality. 
Second, we proposed a novel revision-controllable method that introduces an allowed revision window within the beam search pruning process. 
This approach effectively filters out candidate translations that could lead to extensive revisions, 
thereby significantly reducing flickering and enhancing the stability of the translation.
Moreover, by limiting the maximum number of tokens that can be revised, 
our method successfully prevents extreme instances of instability, 
thereby significantly improving user experience.
Importantly, these improvements were achieved without a significant compromise on translation quality. 

% \jc{1.case study.}
\end{CJK*}

% Below is an example of how to insert images. Delete the ``\vspace'' line,
% uncomment the preceding line ``\centerline...'' and replace ``imageX.ps''
% with a suitable PostScript file name.
% -------------------------------------------------------------------------
% \begin{figure}[htb]
	
% 	\begin{minipage}[b]{1.0\linewidth}
% 		\centering
% 		\centerline{\includegraphics[width=8.5cm]{image1}}
% 		%  \vspace{2.0cm}
% 		\centerline{(a) Result 1}\medskip
% 	\end{minipage}
% 	%
% 	\begin{minipage}[b]{.48\linewidth}
% 		\centering
% 		\centerline{\includegraphics[width=4.0cm]{image3}}
% 		%  \vspace{1.5cm}
% 		\centerline{(b) Results 3}\medskip
% 	\end{minipage}
% 	\hfill
% 	\begin{minipage}[b]{0.48\linewidth}
% 		\centering
% 		\centerline{\includegraphics[width=4.0cm]{image4}}
% 		%  \vspace{1.5cm}
% 		\centerline{(c) Result 4}\medskip
% 	\end{minipage}
% 	%
% 	\caption{Example of placing a figure with experimental results.}
% 	\label{fig:res}
% 	%
% \end{figure}

\bibliographystyle{IEEEbib}
\bibliography{ref}

\end{document}